\pdfoutput=1

\documentclass[11pt]{article}

\usepackage{acl}

\usepackage{times}
\usepackage{latexsym}
\usepackage{graphicx}
\usepackage{amsmath}
\usepackage[T1]{fontenc}
\usepackage{booktabs}
\usepackage{multirow}
\usepackage{multicol}
\usepackage{algorithm}
\usepackage{algorithmic}
\usepackage{graphicx}
\usepackage{subcaption}
\usepackage[utf8]{inputenc}

\usepackage{microtype}

\usepackage{inconsolata}

\usepackage{xcolor}
\definecolor{myp}{RGB}{128, 0, 128}
\newcommand{\name}{\textit{STEP}}
\title{\name: Success-Rate-Aware Trajectory-Efficient Policy Optimization}

\author{
\begin{tabular}{c}
Yuhan Chen$^{1}$ \quad  Yuxuan Liu$^{2}$ \quad  Long Zhang$^{3}$ \quad Pengzhi Gao$^{1}$ \quad Jian Luan$^{1}$\quad Wei Liu$^{1}$\thanks{\ \ Corresponding author.}
\end{tabular}
\\ \vspace{.5mm}
    \begin{tabular}{c}
    $^1$MiLM Plus, Xiaomi Inc.  \quad $^2$Renmin University of China \quad $^3$Wuhan University\\
    \end{tabular}
    \\ \vspace{.5mm}
    \begin{tabular}{c}
    \texttt{\{chenyuhan5, gaopengzhi, luanjian, liuwei40\}@xiaomi.com}\\
    \texttt{yuxuanliu@ruc.edu.cn}\\
    \texttt{zlongooo@whu.edu.cn}\\
    \end{tabular}
    \vspace{2mm} \\
}

\begin{document}
\maketitle
\begin{abstract}
Multi-turn interaction remains challenging for online reinforcement learning. A common solution is trajectory-level optimization, which treats each trajectory as a single training sample. However, this approach can be inefficient and yield misleading learning signals: it applies uniform sampling across tasks regardless of difficulty, penalizes correct intermediate actions in failed trajectories, and incurs high sample-collection costs. To address these issues, we propose \name\ (Success-rate-aware Trajectory-Efficient Policy optimization), a framework that dynamically allocates sampling based on per-task success rates and performs step-level optimization. \name\ maintains a smoothed success-rate record to guide adaptive trajectory resampling, allocating more effort to harder tasks. It then computes success-rate-weighted advantages and decomposes trajectories into step-level samples. Finally, it applies a step-level GRPO augmentation to refine updates for low-success tasks. Experiments on OSWorld and AndroidWorld show that \name\ substantially improves sample efficiency and training stability over trajectory-level GRPO, converging faster and generalizing better under the same sampling budget.
\end{abstract}

\section{Introduction}
\label{sec:intro}
Large language models (LLMs) have been increasingly adopted as agents for multi-turn decision-making, where they must reason, plan, and act over extended interactions with delayed and sparse rewards. Such applications include program synthesis~\cite{Zhang2024CodeAgentEC}, interactive gameplay~\cite{Narasimhan2015LanguageUF}, robotic control~\cite{Brohan2023RT2VM} and GUI automation~\cite{Qin2025UITARSPA, Ye2025MobileAgentv3FA}. To improve these agents’ adaptability, reinforcement learning (RL) has become a key paradigm for online policy optimization through feedback-driven interaction.

\begin{figure}[t]
    \centering
    \includegraphics[width=\linewidth]{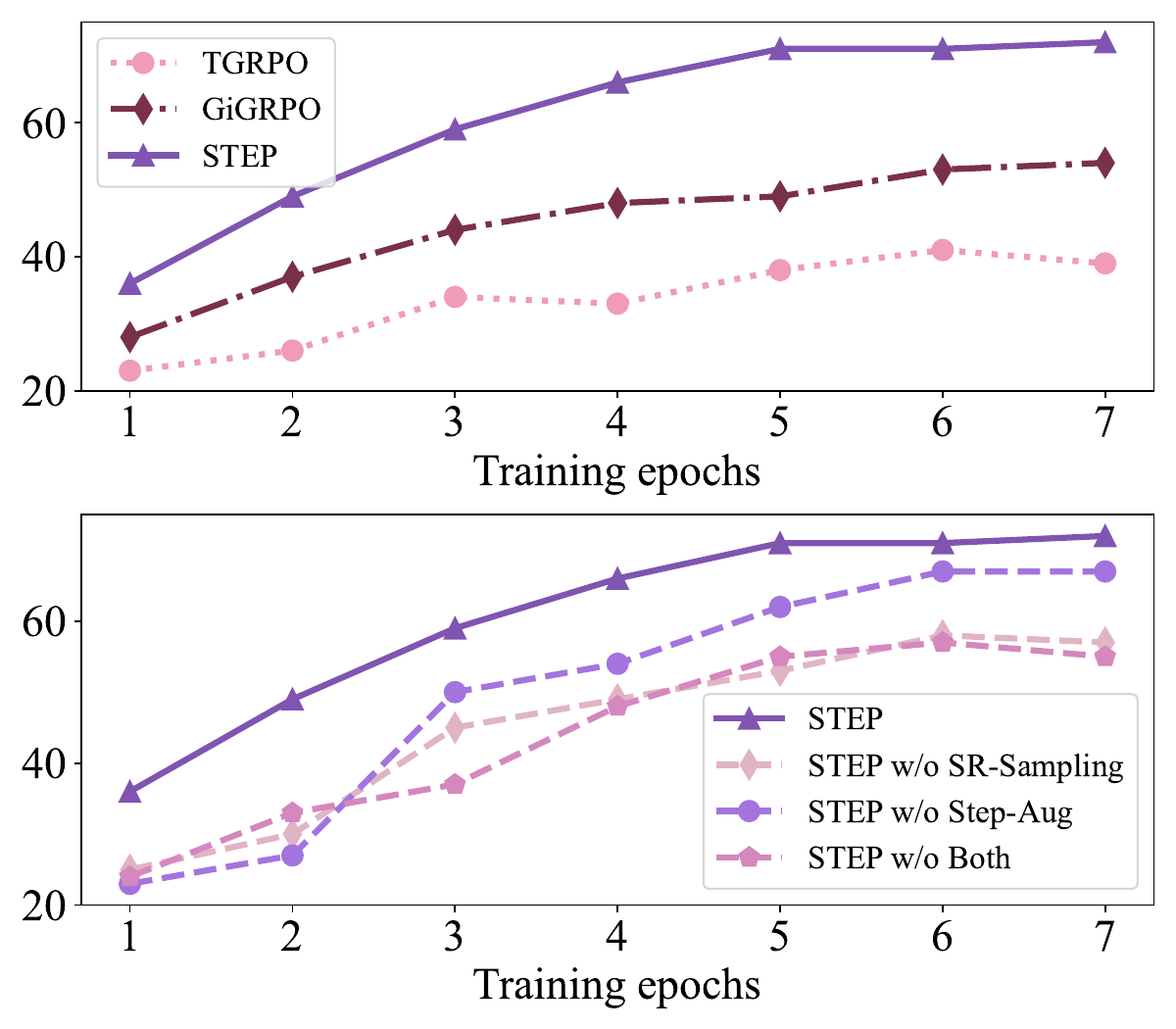}
    \caption{
Number of tasks in the OSWorld training subset (128 tasks) that achieve a success rate above 60\% during training across different methods. We report results for trajectory-level GRPO (T-GRPO), GiGRPO, our proposed method (\name) and three ablation variants of \name. Further details are provided in Section~\ref{sec:result}.}
    \label{fig:sr-compare}
\end{figure}

Among RL-based methods, Group Relative Policy Optimization (GRPO)~\cite{Shao2024DeepSeekMathPT} has been widely adopted for its efficiency and scalability. In multi-turn settings, a common practice is to treat each trajectory—the full sequence of decisions and feedback from an episode—as a single training sample. The policy is then optimized at the trajectory level, updating parameters based on the aggregated return or relative score of the entire episode.

However, this trajectory-level formulation introduces several critical challenges:

 (1) \textbf{Uniform sampling across tasks.} GRPO allocates equal sampling effort to all tasks, regardless of their difficulty or success rate. In multi-turn settings, where task complexity varies widely, this uniform allocation leads to inefficient use of the sampling budget: the agent repeatedly trains on simple, already-solved tasks while complex, low-success tasks—those that provide the most informative learning signals—receive insufficient attention.

 (2) \textbf{Inaccurate credit assignment.} Multi-turn trajectories often contain a mixture of correct and incorrect actions. When the entire trajectory is treated as a single outcome, correct steps within an otherwise failed episode are penalized, propagating inaccurate gradients.

 (3) \textbf{Inefficient sample collection.} Compared with single-turn scenarios, multi-turn tasks require continuous interaction with the environment. Each rollout depends heavily on environment latency, and multiple runs are needed to collect a single trajectory, making sample collection highly inefficient.

To address these challenges, we propose \name (Success-rate-aware Trajectory-Efficient Policy optimization), a framework that dynamically allocates sampling and learning effort based on per-task success rates and performs fine-grained, step-level optimization. \name\ maintains smoothed success-rate records to guide adaptive trajectory resampling, decomposes only the successful trajectories into step-level samples for success-rate-weighted credit assignment, and applies step-level GRPO augmentation for low-success tasks. This design enables both efficient use of sampling budgets and stable, high-quality learning, leading to faster convergence and better generalization in multi-turn RL scenarios.

We evaluate our approach on two general-purpose GUI benchmarks, OSWorld and AndroidWorld, which feature complex, multi-turn interaction environments suitable for comprehensive assessment. As depicted in Figure~\ref{fig:sr-compare}, our \name\ consistently outperforms existing methods, achieving higher efficiency and effectiveness than trajectory-level GRPO.

To sum up, we make three major contributions:

 (1) We systematically analyze the challenges in multi-turn reinforcement learning and provide several key insights.
 
 (2) Based on these insights, we propose \name, a step-level training framework specifically designed for multi-turn reinforcement learning.

 (3) Through extensive experiments, we demonstrate the effectiveness of \name, achieving improvements both in training efficiency and performance.

\section{Related Work}
\label{sec:relate}
\subsection{LLM Agents for Multi-Turn Interaction}
Recent advances in Large Language Models (LLMs)~\cite{Yao2022ReActSR, Brohan2023RT2VM} have expanded their role from static language understanding to interactive agents that perceive, reason, and act in dynamic environments. Research has increasingly explored these agents across diverse domains, including embodied navigation in simulated homes~\cite{Shridhar2020ALFWorldAT, Li2024EmbodiedAI}, multi-step web and mobile task execution leveraging structured pages and APIs~\cite{Hong2023CogAgentAV, Gur2023ARW, Furuta2023MultimodalWN, Gou2024NavigatingTD}, and adaptive decision-making in interactive games~\cite{Narasimhan2015LanguageUF, Wang2024MobileAgentv2MD}. A common goal across these studies is to enable LLMs to maintain coherent perception–reasoning–action loops over multiple turns, which requires robust contextual understanding and long-horizon planning.
Some approaches~\cite{Schick2023ToolformerLM, Wang2023VoyagerAO, Zhang2023AppAgentMA} address this by constructing modular workflows that combine multiple components to perform complex tasks, showing potential for improved performance. More recently, methods have increasingly focused on training LLMs directly on interaction data using supervised fine-tuning (SFT)~\cite{Zhang2023YouOL}, or reinforcement learning (RL)~\cite{Sutton1998ReinforcementLA}, allowing models to acquire task-relevant patterns from environmental interactions.

\subsection{Reinforcement Learning for Large Language Models}
An early and influential application of reinforcement learning (RL) in large language models (LLMs) is RLHF~\cite{Stiennon2020LearningTS, Ouyang2022TrainingLM}, which aligns model outputs with human preferences. More recently, RL has been increasingly employed to enhance reasoning and logical deduction in LLMs, using methods such as PPO~\cite{Schulman2017ProximalPO}, DPO~\cite{Rafailov2023DirectPO}, and GRPO~\cite{Shao2024DeepSeekMathPT}. In particular, group-based RL algorithms such as GRPO, Dr. GRPO~\cite{Liu2025UnderstandingRT}, and DAPO~\cite{Yu2025DAPOAO} have shown promise due to their low computational cost and efficient updates. By leveraging a group of samples from the same query, these methods estimate advantages without introducing an additional value function. They have achieved strong performance in tasks such as mathematical reasoning, search, and tool use, though these tasks are predominantly single-turn.
Recent studies~\cite{Wang2025RAGENUS, Lu2025ARPOEndtoEndPO, Ye2025MobileAgentv3FA} have extended these approaches to multi-turn interactions by treating entire trajectories as sequences of independent steps. This simplification, however, overlooks the fundamental challenges inherent in multi-turn settings. One exception is GiGRPO~\cite{Feng2025GroupinGroupPO}, which introduces a two-level structure to estimate per-step advantages from grouped states across trajectories. Nevertheless, it focuses mainly on long-horizon training due to device constraints and does not systematically address the broader challenges of multi-turn interactions.

In this work, we first present a systematic analysis of these challenges in multi-turn RL. Building on this foundation, we propose a success-rate-guided optimization strategy that reallocates sampling resources and strengthens step-level learning signals, thereby improving both sample efficiency and policy performance of LLM agents in multi-turn interactions.

\section{Preliminaries}
\label{sec:pre}
Formally, given a task $Q$ and an initial environment state, the LLM agent interacts with the environment over multiple steps to accomplish the task.
At each step $t$, the agent observes a state 
$S_t = \langle Q, H_t, I_t \rangle$, 
where $Q$ denotes the task description, $H_t$ the interaction history, and $I_t$ the current environment observation (e.g., a screenshot).  
Based on $S_t$, the agent generates a textual response using an LLM policy $\pi_{\theta}$, from which an action $A_t$ is extracted and executed in the environment, yielding an immediate reward $R_t$.  
This process continues until the episode terminates or the step limit is reached, producing a trajectory:
\[\mathcal{T} = \{\mathcal{S}_1, R_1^{*}, \ldots,\mathcal{S}_t, R_t^{*}, \ldots ,\mathcal{S}_T, R_T\},
\]
where each state-action pair $\mathcal{S}_t := (S_t, A_t)$ corresponds to step $t$. Here, $*$ indicates that intermediate rewards $R_t^*$ may be unavailable in some scenarios, and $R_T$ denotes the final trajectory reward $R_{\mathcal{T}}$.
The training objective then is to update the LLM policy $\pi_{\theta}$ to maximize the expected reward across tasks.

\section{Revisiting Trajectory-Level GRPO}
In this section, we revisit trajectory-level group-based reinforcement learning (T-GRPO), a trajectory-focused extension of GRPO, and discuss the challenges it faces in complex multi-turn tasks.

\subsection{Trajectory-Level GRPO Mechanism}
\label{app:tg}
Trajectory-Level GRPO is a straightforward application of GRPO to multi-turn scenarios.
For a given task, the agent first samples a group of $N$ trajectories 
$\mathcal{G}_{\mathcal{T}} = \{\mathcal{T}_{1}, \ldots, \mathcal{T}_{N}\}$ under the old policy $\pi_{\theta_{\text{old}}}$, with each trajectory $\mathcal{T}$ associated with a final reward $R$. 

The trajectory advantages are then computed based on the reward statistics of the sampled group:
\[
\mathrm{Adv} (\mathcal{T}_{i}) = \frac{ R (\mathcal{T}_{i}) - \text{mean}\big (R (\mathcal{T}_{j}) \mid \mathcal{T}_{j} \in \mathcal{G}_{\mathcal{{T}}}\big) }
       { \text{std}\big (R (\mathcal{T}_{j}) \mid \mathcal{T}_{j} \in \mathcal{G}_{\mathcal{T}}\big) }.
\]
In this formulation, each state $S_t$ within the trajectory contains the concatenation of all previous screenshots and textual responses, forming an accumulation of all past states and actions. Naturally, the trajectory $\mathcal{T}$ is equivalent to the final state-action pair $\mathcal{S}_T$, summarizing the agent’s entire interaction with the environment.

\subsection{Challenges in T-GRPO}
While effective for single-turn tasks, GRPO may encounter several challenges in complex multi-turn reasoning (T-GRPO), including: 
 (1) Uniform sampling across tasks,
 (2) Misaligned learning signals, and 
 (3) Inefficient sample collection.
In the following, we discuss each of these challenges in detail.

\begin{figure}[t]
    \centering
    \includegraphics[width=\linewidth]{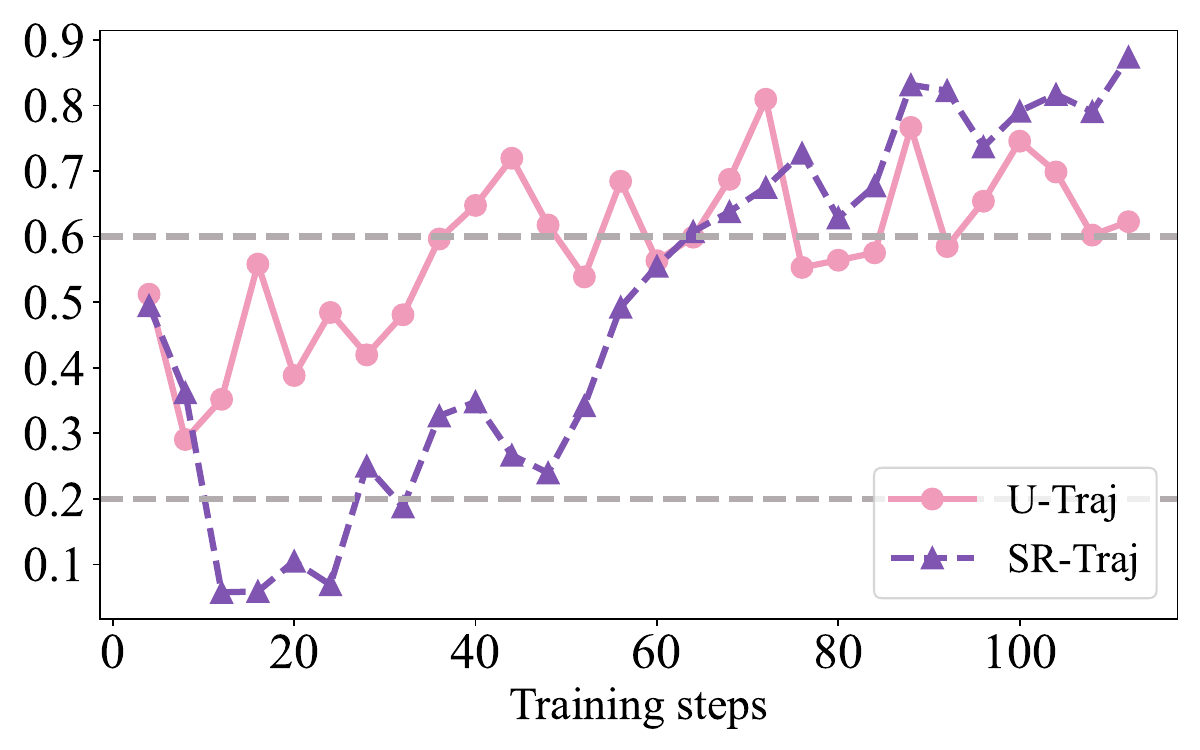}

    \caption{
Proportion of high-success task trajectories over training under different sampling strategies.}
    \label{fig:hs-compare}
 
\end{figure}

\paragraph{Uniform Sampling Across Tasks}
Multi-turn tasks are generally more challenging, often requiring multiple epochs of training to reach satisfactory performance, which highlights the need for efficient sampling strategies. 
However, T-GRPO allocates an equal number of sampled trajectories to each task, without considering differences in task difficulty or success rate. 
This uniform allocation often results in a large portion of successful trajectories coming from simple, already-solved tasks, while challenging tasks—those that could provide more informative learning signals—receive limited attention. 
To systematically quantify this imbalance, we measure the proportion of high-success trajectories (success rate $>$ 80\%) among all successful sampled trajectories and the result is shown in the Figure~\ref{fig:hs-compare}. 
We observe that this ratio remains consistently high throughout training, with the uniform sampling strategy (U-Traj)—even in the early stages—averaging around 60\%. 
However, as shown in Figure~\ref{fig:sr-compare}, the number of high-success tasks is relatively small at the beginning of training. 
This suggests that under the GRPO method, the agent’s exposure to informative learning signals is limited in the early phase, causing it to repeatedly train on tasks it has already mastered, which can trap the model in local optima and reduce its generalization ability.

\paragraph{Misaligned Learning Signals}
T-GRPO assigns rewards solely based on the outcome of a trajectory. 
However, in multi-turn reasoning, a single incorrect step can invalidate the entire trajectory, even if most preceding steps are correct. 
We hypothesize that such coarse reward assignment overlooks valuable intermediate learning signals. 
To investigate this, we randomly sampled 100 failed trajectories from the OSWorld training data and manually compared their step sequences with those of successful trajectories. 
Surprisingly, 78\% of failed trajectories contained sub-sequences identical to those in successful ones, and at the step level, the valid proportion reached 38.56\%. 
Figures in Appendix~\ref{app:case} present representative cases from both OSWorld and AndroidWorld, where correct intermediate reasoning steps are penalized due to incorrect final predictions. 
These results indicate that GRPO’s trajectory-level reward fails to capture partial correctness, limiting its ability to assign credit effectively. 

 \begin{figure}[t]
    \centering
    \begin{subfigure}[b]{\linewidth}
    \includegraphics[width=\linewidth]{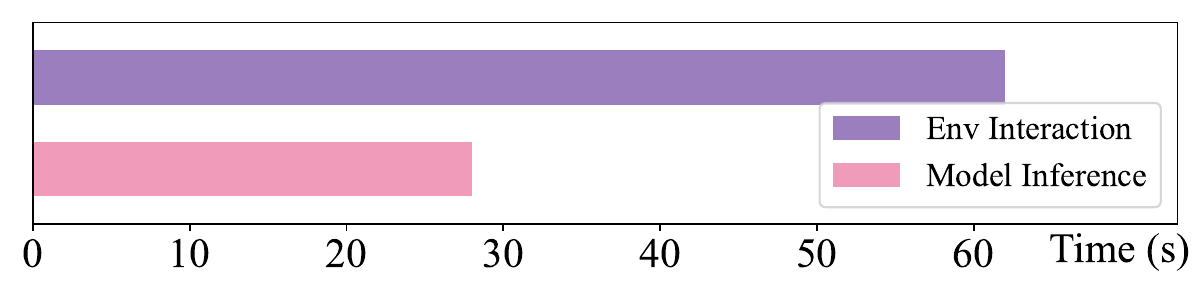}

    \caption{}
    \label{fig:time-compare-a}
    \end{subfigure}
    \begin{subfigure}[b]{\linewidth}
    \includegraphics[width=\linewidth]{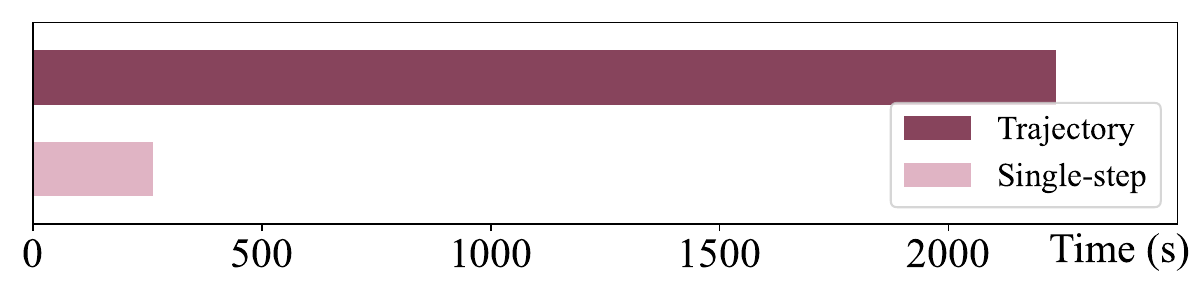}

    \caption{}
    \label{fig:time-compare-b}
    \end{subfigure}
\caption{
Analysis of sample collection efficiency. 
 (a) Wall-clock time per turn (environment vs. model inference). 
 (b) Sampling efficiency: trajectories vs. single steps.
}

\end{figure}

\paragraph{Inefficient Sample Collection}

Another critical challenge is the inefficiency of trajectory-level sampling. 
GRPO requires generating complete multi-turn trajectories for each training update, with each trajectory involving multiple model inferences and environment interactions. 
To quantify this, we measure the wall-clock time of a batch rollout of 256 trajectories and report the time per turn (Figure~\ref{fig:time-compare-a}). The results indicate that most of the time is spent on environment interactions, while model inference accounts for only a small fraction.
We further compare trajectory-level and single-step sampling. To ensure a fair comparison, all turns from the trajectory-level rollouts (average trajectory length is around 15, total turns are 3,856) are re-inferred in parallel to collect timing statistics for single-step sampling.
As shown in Figure~\ref{fig:time-compare-b} (b), collecting complete multi-turn trajectories is far less efficient than parallel single-step rollouts—about 8.5× slower. This bottleneck arises primarily from the high cost of environment interactions and the sequential nature of trajectory rollouts, which limits parallel inference.

From these observations, we derive the following insights:

 (1) Uniform sampling wastes a significant portion of the sampling budget on already mastered tasks, hindering the learning of other valuable tasks. Sampling budgets should therefore be dynamically adjusted based on per-task success rates.

 (2) Failed trajectories produce a large amount of misleading learning signals; focusing on successful trajectories can more effectively guide the model toward correct behaviors.

 (3) Increasing the number of samples through parallelization—without incurring additional interaction costs with the environment—can significantly enhance sampling efficiency.

\begin{figure*}[t]
    \centering
    \includegraphics[width=\linewidth]{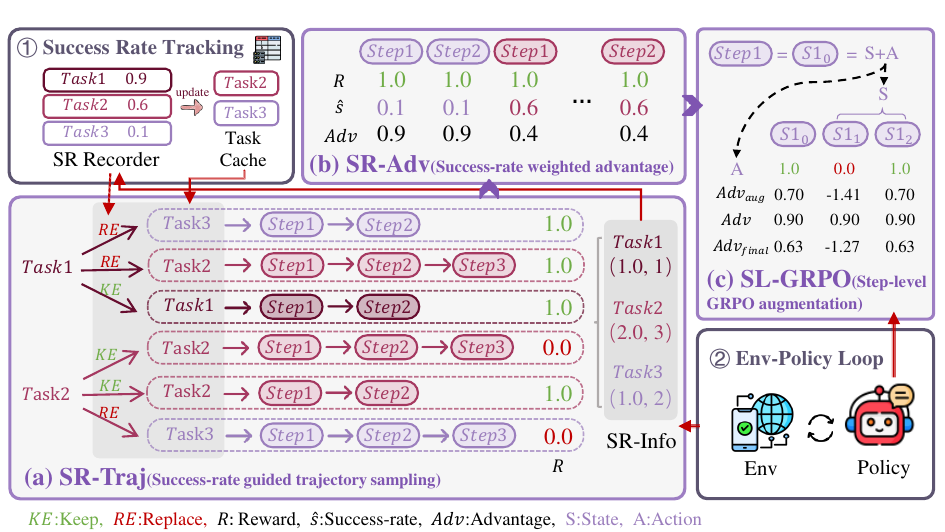}

\caption{
The framework of our \name, which consists of three core components: 
 (a) Success-rate guided trajectory sampling (SR-Traj, \S\ref{md:sr-sample}), which allocates the sampling budget based on dynamically updated success rates; 
 (b) Success-rate weighted advantage (SR-Adv, \S\ref{md:sr-adv}), where we design a specific advantage function for \name; and 
 (c) Step-level GRPO augmentation (SL-GRPO, \S\ref{md:step-aug}), which performs data augmentation without additional environment interaction costs.
}
    \label{fig:astar}
    
\end{figure*}

\section{Method}
Building upon the insights above, we introduce our method  \name\ (Success-rate-aware Trajectory-Efficient Policy optimization). As shown in the Figure~\ref{fig:astar},  \name\ adaptively adjusts both trajectory sampling and policy learning based on per-task success rates through three core components: (a) Success-rate guided trajectory sampling (SR-Traj), (b) Success-rate weighted advantage (SR-Adv), and (c) Step-level GRPO augmentation (SL-GRPO).

\subsection{Success-Rate Guided Trajectory Sampling}
\label{md:sr-sample}

We propose a success-rate guided trajectory sampling strategy that dynamically reallocates sampling resources based on per-task success rates.

To track success rates, we maintain two key structures:
\begin{itemize}
    \item \textbf{Global success-rate record} $\hat{s}$, which stores the estimated success rate $\hat{s}_i$ for each task $Q_i$.
    \item \textbf{Task cache} $C_Q$, containing tasks with intermediate success rates ($0 < \hat{s}_i < s_0$), where $s_0$ is a predefined threshold.
\end{itemize}
Based on these structures, the sampling procedure is organized into two main parts: (i) Sampling Budget Reallocation and (ii) Tracking Update.

\paragraph{Sampling Budget Reallocation} 
In each trajectory collection round, every task $Q_i$ is expanded into $N$ copies (following GRPO). 
For each copy\footnote{To preserve diversity and avoid forgetting, one copy is kept; replacement is applied to the remaining $N$-1 copies.}, a \textbf{replacement function} decides whether to substitute the original task with one sampled from $C_Q$. The replacement probability is defined by a logistic function:
\[
p_{\text{rep}} (\hat{s}_i) = \frac{1}{1 + \exp\big (-\kappa (\hat{s}_i - s_0)\big)},
\]
where $\kappa$ controls the sharpness of the transition around $s_0$. So the final $\hat{Q}_{i,j}$ for the $j$-th copy of task $Q_i$ is then
\[
\begin{aligned}
& \text{rep}_{i,j}  \sim \mathrm{Bernoulli}\big (p_{\text{rep}} (\hat{s}_i)\big), \\
\hat{Q}_{i,j} &=
\begin{cases}
Q_k, & \text{if } \text{rep}_{i,j}=1,~Q_k \in C_Q,\\
Q_i, & \text{otherwise.}
\end{cases}
\end{aligned}
\] 
After performing this sampling reallocation, we proceed to collect the trajectories. Intuitively, tasks with higher success rates are more likely to be replaced, thereby focusing sampling on tasks with lower success rates, which are expected to provide more informative signals.

\paragraph{Tracking Update} 
In the end of each collection round, the success-rate record and cache are updated using a smoothed rule to stabilize estimates across varying sample sizes. Let $N_i^{\tau}$ and $U_i^{\tau}$ denote the number of collected and successful trajectories for task $i$ in the current round $\tau$. The update is
\[
\begin{aligned}
\hat{s}_i^{\tau} &=
\frac{U_i^{\tau} + \alpha U_i^{\tau-1}}{N_i^{\tau} + \alpha N_i^{\tau-1} } 
= \frac{U_i^{\tau} + \alpha \hat{s}_i^{\tau-1} N}{N_i^{\tau} + \alpha N},\\
\alpha  &= 
\begin{cases}
 1 - \frac{N_i^{\tau}}{N}, &  \text{if } N_i^{\tau} < N, \\
 0, & \text{otherwise},
\end{cases}\\
&C_Q^{\tau} = \{Q_j \mid 0 < \hat{s}_j^{\tau} < s_0\}.
\end{aligned}
\]
Here, we use an adaptive discount factor 
$\alpha$ that adjusts the influence of past estimates according to the number of collected trajectories. This design enables smooth and stable updates across rounds, preventing overly aggressive or unstable changes when data are scarce. The updated $\hat{s}$ and $C_Q$ are carried over to the next round for continued tracking and sampling.

\subsection{Success-Rate-Weighted Advantage}
\label{md:sr-adv}
After trajectory collection, we compute advantages for policy optimization. To reduce the influence of noisy or misleading signals from failed trajectories, we \textbf{use only successful trajectories}. We propose a success-rate-weighted advantage estimator that incorporates trajectory-level information and decomposes trajectories into individual steps to provide fine-grained learning signals.

\paragraph{Success-Rate-Weighted Trajectory Advantage}
Due to the adaptive sampling strategy introduced above, different tasks yield unequal numbers of trajectories, which makes standard group-based advantage normalization unreliable.  Noting that the mean reward of a task group can serve as a proxy for its success rate, we introduce a success-rate-weighted advantage, which combines each trajectory’s reward with the current task success rate.
For each successful trajectory $\mathcal{T}_{i,j}$ of task $Q_i$, the advantage is defined as
\[
\text{Adv} (\mathcal{T}_{i,j}) = (1 - \hat{s}_i) \cdot R_{\mathcal{T}_{i,j}}, \quad R_{\mathcal{T}_{i,j}} > 0,
\]
where $\hat{s}_i$ denotes the smoothed success rate of task $i$, and $R_{\mathcal{T}_{i,j}}$ is the corresponding trajectory reward. This formulation assigns stronger learning signals to tasks with lower success rates, thereby encouraging the model to prioritize underperforming tasks.

\paragraph{Step-level Decomposition}
Each trajectory is then decomposed into step-level samples, with the same advantage assigned to all steps:
\[
\text{Adv} (\mathcal{S}_{\mathcal{T}, t}) = \text{Adv} (\mathcal{T}), \quad \forall t \in [1, T],
\]
where $\mathcal{S}_{\mathcal{T}, t}$ represents the sample at step $t$ of the trajectory, and $T$ is the trajectory length. Since training is performed on step-level samples, this decomposition also allows flexible organization of the history $H_t$ within each $\mathcal{S}_{\mathcal{T}, t}$ (see Section~\ref{sec:pre}) using $t_r$ past responses and $t_I$ past observations.

\subsection{Step-level GRPO Augmentation}  
\label{md:step-aug}
To further improve learning efficiency and stability, we enrich the training set with valuable samples without extra environment interactions and minimal computational cost. Specifically, we selectively perform data augmentation on step samples from tasks \textbf{with low success rates} ($\hat{s}_i <= s_{\text{low}}$), 
which typically correspond to steps with higher trajectory advantages. 

For step samples selected from low-success trajectories, each is represented as
\[
\mathcal{S} = \langle S, A, \text{Adv} (\mathcal{S}) \rangle,
\]
where $S$ denotes the state, $A$ the action, and $\text{Adv} (\mathcal{S})$ the success-rate–weighted advantage.  

To augment these samples, we expand each $\mathcal{S}$ into a set of step-level variants by prompting the model with the same state $S$ to generate $n = N/2 - 1$ alternative actions, where $N$ is the group number we used in trajectory sampling; this setting balances diversity and efficiency without introducing additional hyperparameters.
This \textbf{yields a group of candidate step samples}:
\[
\mathcal{G}_{\mathcal{S}} = \{\mathcal{S}_k = \langle S, A_k \rangle \mid k = 0, \dots, n,~A_0 = A\}.
\]
The original step $\mathcal{S}$ is thus replaced by its augmented group $\mathcal{G}_{\mathcal{S}}$, which represents localized perturbations around the original decision.

For each $\mathcal{S}_k \in \mathcal{G}_{\mathcal{S}}$, we define a step reward based on whether its action $A_k$ \textbf{matches the reference action $A_0$}.  
Each augmented group is then evaluated to assign relative advantages among its members:
\[
\begin{aligned}
R (\mathcal{S}_k) &=
\begin{cases}
1, & \text{if } A_k \text{ matches } A_0, \\
0, & \text{otherwise},
\end{cases}\\[4pt]
\text{Adv}_{\text{aug}} (\mathcal{S}_k) 
&= \frac{ R (\mathcal{S}_k) - \text{mean}\big (R (\mathcal{S}_j) \mid \mathcal{S}_j \in \mathcal{G}_{\mathcal{S}}\big) }
       { \text{std}\big (R (\mathcal{S}_j) \mid \mathcal{S}_j \in \mathcal{G}_{\mathcal{S}}\big) }.
\end{aligned}
\]

Finally, the step-level advantage used for policy optimization combines the trajectory-level credit and the local augmentation signal:
\[
\text{Adv}_{\text{final}} (\mathcal{S}_k) = \text{Adv} (\mathcal{S}) \cdot \text{Adv}_{\text{aug}} (\mathcal{S}_k).
\]

This step-level formulation encourages the policy to refine local action boundaries around high-value steps while maintaining consistency with the trajectory-level objective.

\begin{table*}[t]
\centering
\resizebox{0.95\linewidth}{!}{
\begin{tabular}{lcccc}
\toprule
Method & OSWorld (\textit{Train Set}) & OSWorld & AndroidWorld (\textit{Train Set}) & AndroidWorld \\ \midrule
UI-Tars-DPO-7B   & 41.4 & 16.8  & 29.8 & 29.7\\
+ T-GRPO  &  48.4 (+7.0)  & 18.9 (+2.1) &  33.3 (+3.5) & 31.0 (+1.3)\\
 + GiGRPO & 55.4 (+14.0)   & 21.1 (+4.3)   & 39.2 (+9.4)    &  34.0 (+4.3)   \\
\midrule
+ \name\ (Ours) & \textbf{62.5 (+21.1)}  & \textbf{23.8 (+7.0)} & \textbf{47.6 (+17.8)} & \textbf{45.7 (+16.0)} \\
\bottomrule
\end{tabular}}
\caption{Results on OSWorld and AndroidWorld. The leading results are highlighted with \textbf{bold fonts}. Our \name\ demonstrates superior performance in both benchmarks.} 
\label{tab:main}
\end{table*}

\section{Experiments}
\subsection{Experiments Setups}
\label{sec:setting}
\paragraph{Benchmarks.} 

We selected two widely recognized benchmarks in the graphical user interface (GUI) domain—\textbf{OSWorld}~\cite{Xie2024OSWorldBM} and \textbf{AndroidWorld}~\cite{Rawles2024AndroidWorldAD}. These benchmarks were chosen because they represent classical and challenging multi-turn interaction tasks in GUI-based environments, making them suitable for evaluating the robustness and generalization ability of our method.  \textbf{OSWorld} provides a real-computer environment containing 369 tasks across diverse domains such as office productivity, web browsing, system management, and multi-application workflows. Following ARPO~\cite{Lu2025ARPOEndtoEndPO}, we sample 128 tasks from the OSWorld benchmark as our training set.  
\textbf{AndroidWorld} offers a fully functional Android environment with reward signals for 116 programmatic tasks across 20 real-world Android applications. We select a subset of 43 tasks from these applications as the training set.  
Both benchmarks use rule-based rewards evaluated only after completing a trajectory, assigning 1.0 to successful executions and 0.0 otherwise.

\paragraph{Baselines and Training Details}
We adopt UI-Tars-DPO-7B~\cite{Qin2025UITARSPA} as our base model, and select (1)trajectory-level GRPO (T-GRPO)~\cite{Shao2024DeepSeekMathPT} and (2)GiGRPO\cite{Feng2025GroupinGroupPO}\footnote{In the complex GUI scenario, step-level state clustering in GiGRPO is not suitable. In this context, GiGRPO denotes a variant of GRPO in which trajectories are split into step samples, with the trajectory-level advantage distributed to each step for training.} as our baselines. Methods such as DAPO~\cite{Yu2025DAPOAO} or other replay-based approaches (e.g., ARPO) are excluded, as they can be integrated with our approach.  
All RL training methods use identical hyperparameters. The training batch size is 16, with a rollout number \(N = 16\), and a PPO train size of 256. The temperature is 0.7 during training and 0 during evaluation. We apply dynamic batch updates in our method and GiGRPO, as both decompose trajectories into step-level sample, resulting in a variable number of step samples. For each step sample, the number of history responses \(t_r\) and screenshots \(t_I\) in \(H_t\) are 3 and 0, respectively. 
The specific hyperparameters of our method are: threshold \(s_0 = 0.6\), low threshold \(s_{\text{low}} = 0.2\), and \(\kappa = 10\). Full training settings and hyperparameters are provided in Appendix~\ref{app:setting}.

\begin{table}[t]
\centering
\resizebox{\linewidth}{!}{
\begin{tabular}{lcc}
\toprule
Method  &  OSWorld (\textit{Train Set}) & OSWorld \\\midrule
\name\ (Ours) & \textbf{62.5}  & \textbf{23.8} \\\midrule
w/o SR-Sampling & 57.0 (-5.5) &  21.7 (-1.1)   \\
w/o Step-Aug & 60.1 (-2.4) & 23.0 (-0.8)   \\
w/o Both & 56.2 (-6.3) & 21.4 (-1.4) \\
\bottomrule
\end{tabular}}
\caption{The results of the ablation study.} 
\label{tab:ab}
\end{table}

\subsection{Results}
\label{sec:result}
The overall performance on OSWorld and AndroidWorld is summarized in Table~\ref{tab:main}, while the ablation results for \name's core components are shown in Table~\ref{tab:ab}. Additionally, Figure~\ref{fig:sr-compare} illustrates the evolution of task categories with success rates above 60\%  across training epochs, offering further insights into the learning dynamics.

\subsubsection{Main Results}  
As shown in Table~\ref{tab:main}, \name\ consistently outperforms all baselines across both benchmarks, demonstrating clear advantages on the \textit{Train Set} and overall evaluations. On OSWorld (\textit{Train Set}), \name\ reaches 62.5, exceeding T-GRPO and GiGRPO by 14.1 and 7.1 points, respectively. On AndroidWorld, it achieves 45.7 overall, surpassing T-GRPO by 14.7 and GiGRPO by 11.7 points, confirming its robustness across domains.

Figure~\ref{fig:sr-compare} further illustrates that GiGRPO and \name, both using step-level samples, improve faster and achieve higher scores than T-GRPO. We hypothesize that step-level training helps reorganize trajectories and maintain shorter context windows, exposing the model to more state-similar samples and accelerating convergence. Moreover, compared with GiGRPO, \name\ performs even better, likely due to its dynamic adjustment of in-batch training samples based on success rates. This adaptive sampling strategy encourages the model to focus on under-learned or unstable tasks, avoiding overfitting and reducing the risk of local optima. Our step-level augmentation further enriches training data by generating fine-grained samples from challenging tasks, providing higher-quality learning signals and faster convergence.

It is worth noting that the improvement from the \textit{Train Set} to overall evaluation is less consistent on OSWorld than on AndroidWorld. On OSWorld, \name\ improves by 21.1 points on the \textit{Train Set} and by 7.0 points overall, whereas on AndroidWorld, the improvements are 17.8 and 16.0 points, respectively. This indicates that improvements observed during OSWorld training do not fully transfer to the overall evaluation.
We hypothesize that this discrepancy stems from the OSWorld \textit{Train Set} being constructed via pre-sampling~\cite{Lu2025ARPOEndtoEndPO}, which selects tasks more likely to yield rewards. As a result, the remaining tasks in the overall set are harder and less aligned with the training distribution, leading to weaker generalization. In contrast, the AndroidWorld \textit{Train Set} was randomly sampled, ensuring each app contributes at least one task, which allows overall performance to better reflect \textit{Train Set} gains.

\subsubsection{Ablation Study}  
We conduct ablation experiments to assess the contribution of each core component in \name. Three variants are evaluated:
 (1) w/o SR-Sampling, which removes success-rate sampling;
 (2) w/o Step-Aug, which removes step augmentation;
 (3) w/o Both, which disables both components, leaving only our advantage estimation applied to GiGRPO.

As shown in Table~\ref{tab:ab}, removing either SR-Sampling or Step-Aug results in a clear performance drop, demonstrating that both components are essential to the effectiveness of \name. Specifically, excluding SR-Sampling causes a 7.1-point decrease on the \textit{Train Set}, suggesting that adaptive sampling plays a key role in stabilizing training and improving sample quality. Meanwhile, as shown in Figure~\ref{fig:sr-compare}, removing Step-Aug slows down the growth in task diversity, indicating that augmenting intermediate steps provides additional supervision signals that enhance learning efficiency. The variant without both components (21.4 on OSWorld) performs slightly better than GiGRPO (23.8). This result suggests that our advantage estimation is reasonable and that training solely on successful trajectories indeed provides measurable benefits.

\section{Discussion}
\subsection{Mitigating Over-Sampling of Mastered Tasks}
\label{sec:res-vs}

 Uniform task sampling (U-Traj, used in T-GRPO and GiGRPO) tends to over-train on high-success tasks. In contrast, our method, Success-Rate Guided Trajectory Sampling (SR-Traj), addresses this imbalance by adjusting the sampling budget based on the dynamic task success rates.
To evaluate this effect, we track the proportion of trajectories from high-success tasks ($\hat{s_i} \ge 0.8$ ) during training, computing the average proportion every four training steps. Figure~\ref{fig:hs-compare} visualizes these trends, providing a direct comparison of how U-Traj and SR-Traj handle mastered tasks over time.

\paragraph{Results} 
As shown in Figure~\ref{fig:hs-compare}, SR-Traj substantially reduces the proportion of high-success trajectories early in training (below 0.2), providing more opportunities to explore less-mastered tasks and generating a richer set of informative samples. As training progresses, this proportion increases, reflecting the overall improvement in task success rates. This dynamic pattern indicates that, given the same sampling budget, SR-Traj effectively mitigates premature overfitting to easier tasks, supporting more efficient learning and robust generalization in later stages.

\begin{figure}[t]
    \centering
    \includegraphics[width=\linewidth]{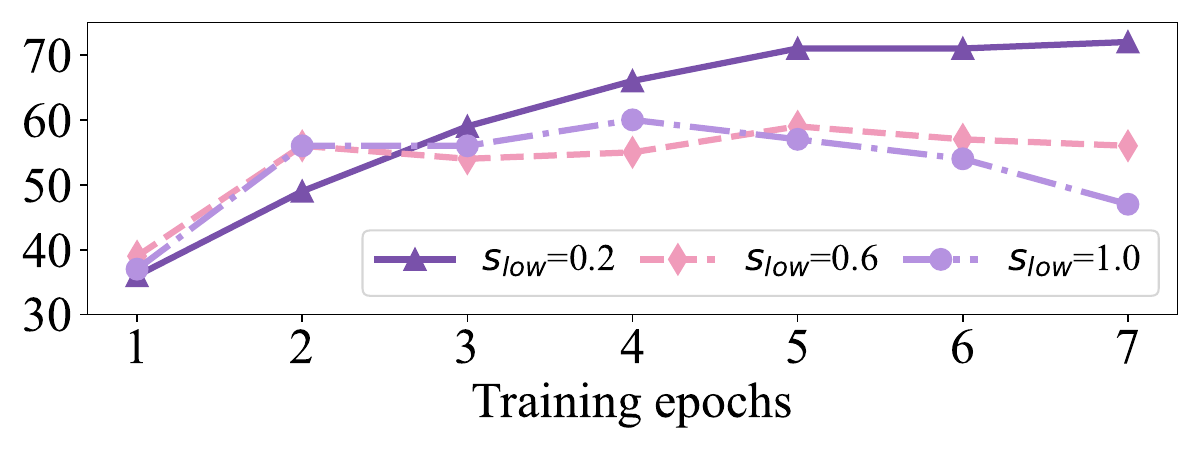}

    \caption{ Number of tasks in OSWorld training sub-
set (128 tasks) with a success rate above 60\% during
training across different $s_{low}$ based on our \name.}
    \label{fig:sa-th}
\end{figure}

\subsection{Effect of Success-Rate Threshold on Step Augmentation}
\label{sec:aug-radio}
As mentioned previously, we applied step augmentation to tasks with low success rates ($s_{low}$ = 0.2). To further verify the effect of this design, we conducted experiments by applying augmentation to tasks under different success-rate thresholds (0.2, 0.6 and 1.0). We track the evolution of task categories with success rates above 60\% across training epochs according to the main experiments. The results are present in Figure~\ref{fig:sa-th}.
\paragraph{Results} 
As illustrated in the figure, all settings exhibit a rapid improvement in the early stages of training, confirming that step augmentation serves as an effective approach to accelerate model learning. However, the growth trends for higher thresholds (both 0.6 and 1.0) slow down noticeably and even decline in later stages, suggesting that applying augmentation to a broader range of tasks may suppress the learning signals from low-success tasks, thereby slightly reducing the model’s generalization ability.

\begin{table}[t]
\centering
\resizebox{0.8\linewidth}{!}{
\begin{tabular}{lcc}
\toprule
Method & Time (min/step) & Speedup \\
\midrule
T-GRPO & 45.67 & 1.0× \\
GiGRPO & 24.59   & 1.86× \\
Our \name\ & 26.25 & 1.74×  \\
\bottomrule
\end{tabular}}
\caption{Average training time per step for different methods.}
\label{tab:eff}
\end{table}

\subsection{Training Efficiency}
\label{sec:efficient}
We evaluate training efficiency by measuring the average time per training step.  The efficiency experiments are conducted on OSWorld, where we deploy a unified setup of 16 GPUs across two nodes. The environments are simulated on a remote server with 128 parallel instances. 
\paragraph{Results} 
Table~\ref{tab:eff} summarizes the results. Both GiGRPO and our \name\ substantially speed up training, achieving 1.86× and 1.74× faster steps, respectively, nearly halving the time compared to T-GRPO. We attribute this improvement to the substantial shortening of context length in both methods, which reduces inference and training overhead.  Remarkably, our method retains high efficiency despite the extra cost of step augmentation during sampling. This is likely due to the avoidance of environment interaction overhead and the efficient parallel inference enabled by step rollout.

\section{Conclusion}
In this paper, we systematically analyze the challenges of multi-turn reinforcement learning, including uniform task sampling, inaccurate credit assignment, and inefficient sample collection. To address these issues, we propose \name\ (Success-rate-aware Trajectory-Efficient Policy Optimization), which maintains smoothed per-task success rates to guide adaptive trajectory resampling, decomposes trajectories into step-level samples with success-rate-weighted advantages, and applies step-level GRPO augmentation to improve learning on low-success tasks. Extensive experiments demonstrate that \name\ substantially outperforms trajectory-level GRPO in both efficiency and effectiveness. As agents face increasingly complex environments, reinforcement learning methods must evolve accordingly. For multi-turn scenarios, we envision \name\ as a step-level framework that can serve as a foundation and reference point, offering insights for future research in efficient and adaptive multi-turn RL.

\section*{Limitations}
Our method improves efficiency and performance compared to GRPO by filtering out failed trajectories and distributing trajectory rewards across all steps of successful ones. However, since failed trajectories are entirely discarded, potentially valuable sub-trajectories within them remain unused, leading to a waste of sampling resources. Moreover, even successful trajectories may contain suboptimal or ineffective actions. These observations indicate that the current reward assignment in multi-turn scenarios remains coarse-grained. Future work could explore more fine-grained, step-level reward mechanisms that selectively leverage informative segments from both successful and partially successful trajectories to further enhance learning stability and accuracy.

\bibliography{anthology,custom}

@article{Lu2025ARPOEndtoEndPO,
  title={ARPO:End-to-End Policy Optimization for GUI Agents with Experience Replay},
  author={Fanbin Lu and Zhisheng Zhong and Shu Liu and Chi-Wing Fu and Jiaya Jia},
  journal={ArXiv},
  year={2025},
  volume={abs/2505.16282},
  url={https://api.semanticscholar.org/CorpusID:278789371}
}

@article{Xie2024OSWorldBM,
  title={OSWorld: Benchmarking Multimodal Agents for Open-Ended Tasks in Real Computer Environments},
  author={Tianbao Xie and Danyang Zhang and Jixuan Chen and Xiaochuan Li and Siheng Zhao and Ruisheng Cao and Toh Jing Hua and Zhoujun Cheng and Dongchan Shin and Fangyu Lei and Yitao Liu and Yiheng Xu and Shuyan Zhou and Silvio Savarese and Caiming Xiong and Victor Zhong and Tao Yu},
  journal={ArXiv},
  year={2024},
  volume={abs/2404.07972},
  url={https://api.semanticscholar.org/CorpusID:269042918}
}

@article{Wang2024MobileAgentv2MD,
  title={Mobile-Agent-v2: Mobile Device Operation Assistant with Effective Navigation via Multi-Agent Collaboration},
  author={Junyang Wang and Haiyang Xu and Haitao Jia and Xi Zhang and Ming Yan and Weizhou Shen and Ji Zhang and Fei Huang and Jitao Sang},
  journal={ArXiv},
  year={2024},
  volume={abs/2406.01014},
  url={https://api.semanticscholar.org/CorpusID:270214232}
}

@article{Gou2024NavigatingTD,
  title={Navigating the Digital World as Humans Do: Universal Visual Grounding for GUI Agents},
  author={Boyu Gou and Ruohan Wang and Boyuan Zheng and Yanan Xie and Cheng Chang and Yiheng Shu and Huan Sun and Yu Su},
  journal={ArXiv},
  year={2024},
  volume={abs/2410.05243},
  url={https://api.semanticscholar.org/CorpusID:273186286}
}

@article{Sutton1998ReinforcementLA,
  title={Reinforcement Learning: An Introduction},
  author={Richard S. Sutton and Andrew G. Barto},
  journal={IEEE Trans. Neural Networks},
  year={1998},
  volume={9},
  pages={1054-1054},
  url={https://api.semanticscholar.org/CorpusID:60035920}
}

@article{Qin2025UITARSPA,
  title={UI-TARS: Pioneering Automated GUI Interaction with Native Agents},
  author={Yujia Qin and Yining Ye and Junjie Fang and Haoming Wang and Shihao Liang and Shizuo Tian and Junda Zhang and Jiahao Li and Yunxin Li and Shijue Huang and Wanjun Zhong and Kuanye Li and Jiale Yang and Yu Miao and Woyu Lin and Longxiang Liu and Xu Jiang and Qianli Ma and Jingyu Li and Xiaojun Xiao and Kai Cai and Chuang Li and Yaowei Zheng and Chaolin Jin and Chen Li and Xiao Zhou and Minchao Wang and Haolin Chen and Zhaojian Li and Haihua Yang and Hai-Yi Liu and Feng Lin and Tao Peng and Xin Liu and Guang Shi},
  journal={ArXiv},
  year={2025},
  volume={abs/2501.12326},
  url={https://api.semanticscholar.org/CorpusID:275788034}
}

@article{Zhang2023YouOL,
  title={You Only Look at Screens: Multimodal Chain-of-Action Agents},
  author={Zhuosheng Zhang and Aston Zhang},
  journal={ArXiv},
  year={2023},
  volume={abs/2309.11436},
  url={https://api.semanticscholar.org/CorpusID:262053313}
}

@article{Ye2025MobileAgentv3FA,
  title={Mobile-Agent-v3: Fundamental Agents for GUI Automation},
  author={Jiabo Ye and Xi Zhang and Haiyang Xu and Haowei Liu and Junyang Wang and Zhaoqing Zhu and Ziwei Zheng and Feiyu Gao and Junjie Cao and Zhengxi Lu and Jitong Liao and Qi Zheng and Fei Huang and Jingren Zhou and Ming Yan},
  journal={ArXiv},
  year={2025},
  volume={abs/2508.15144},
  url={https://api.semanticscholar.org/CorpusID:280699844}
}

@article{Feng2025GroupinGroupPO,
  title={Group-in-Group Policy Optimization for LLM Agent Training},
  author={Lang Feng and Zhenghai Xue and Tingcong Liu and Bo An},
  journal={ArXiv},
  year={2025},
  volume={abs/2505.10978},
  url={https://api.semanticscholar.org/CorpusID:278715074}
}

@article{Wang2025RAGENUS,
  title={RAGEN: Understanding Self-Evolution in LLM Agents via Multi-Turn Reinforcement Learning},
  author={Zihan Wang and Kangrui Wang and Qineng Wang and Pingyue Zhang and Linjie Li and Zhengyuan Yang and Kefan Yu and Minh Nhat Nguyen and Licheng Liu and Eli Gottlieb and Monica Lam and Yiping Lu and Kyunghyun Cho and Jiajun Wu and Fei-Fei Li and Lijuan Wang and Yejin Choi and Manling Li},
  journal={ArXiv},
  year={2025},
  volume={abs/2504.20073},
  url={https://api.semanticscholar.org/CorpusID:278170861}
}

@article{Schulman2017ProximalPO,
  title={Proximal Policy Optimization Algorithms},
  author={John Schulman and Filip Wolski and Prafulla Dhariwal and Alec Radford and Oleg Klimov},
  journal={ArXiv},
  year={2017},
  volume={abs/1707.06347},
  url={https://api.semanticscholar.org/CorpusID:28695052}
}

@article{Ouyang2022TrainingLM,
  title={Training language models to follow instructions with human feedback},
  author={Long Ouyang and Jeff Wu and Xu Jiang and Diogo Almeida and Carroll L. Wainwright and Pamela Mishkin and Chong Zhang and Sandhini Agarwal and Katarina Slama and Alex Ray and John Schulman and Jacob Hilton and Fraser Kelton and Luke E. Miller and Maddie Simens and Amanda Askell and Peter Welinder and Paul Francis Christiano and Jan Leike and Ryan J. Lowe},
  journal={ArXiv},
  year={2022},
  volume={abs/2203.02155},
  url={https://api.semanticscholar.org/CorpusID:246426909}
}

@article{Stiennon2020LearningTS,
  title={Learning to summarize from human feedback},
  author={Nisan Stiennon and Long Ouyang and Jeff Wu and Daniel M. Ziegler and Ryan J. Lowe and Chelsea Voss and Alec Radford and Dario Amodei and Paul Christiano},
  journal={ArXiv},
  year={2020},
  volume={abs/2009.01325},
  url={https://api.semanticscholar.org/CorpusID:221665105}
}

@article{Shao2024DeepSeekMathPT,
  title={DeepSeekMath: Pushing the Limits of Mathematical Reasoning in Open Language Models},
  author={Zhihong Shao and Peiyi Wang and Qihao Zhu and Runxin Xu and Jun-Mei Song and Mingchuan Zhang and Y. K. Li and Yu Wu and Daya Guo},
  journal={ArXiv},
  year={2024},
  volume={abs/2402.03300},
  url={https://api.semanticscholar.org/CorpusID:267412607}
}

@article{Rafailov2023DirectPO,
  title={Direct Preference Optimization: Your Language Model is Secretly a Reward Model},
  author={Rafael Rafailov and Archit Sharma and Eric Mitchell and Stefano Ermon and Christopher D. Manning and Chelsea Finn},
  journal={ArXiv},
  year={2023},
  volume={abs/2305.18290},
  url={https://api.semanticscholar.org/CorpusID:258959321}
}

@article{Liu2025UnderstandingRT,
  title={Understanding R1-Zero-Like Training: A Critical Perspective},
  author={Zi-Yan Liu and Changyu Chen and Wenjun Li and Penghui Qi and Tianyu Pang and Chao Du and Wee Sun Lee and Min Lin},
  journal={ArXiv},
  year={2025},
  volume={abs/2503.20783},
  url={https://api.semanticscholar.org/CorpusID:277322777}
}

@article{Yu2025DAPOAO,
  title={DAPO: An Open-Source LLM Reinforcement Learning System at Scale},
  author={Qiying Yu and Zheng Zhang and Ruofei Zhu and Yufeng Yuan and Xiaochen Zuo and Yu Yue and Tiantian Fan and Gaohong Liu and Lingjun Liu and Xin Liu and Haibin Lin and Zhiqi Lin and Bole Ma and Guangming Sheng and Yuxuan Tong and Chi Zhang and Mofan Zhang and Wang Zhang and Hang Zhu and Jinhua Zhu and Jiaze Chen and Jiangjie Chen and Chengyi Wang and Honglin Yu and Weinan Dai and Yuxuan Song and Xiang Wei and Haodong Zhou and Jingjing Liu and Wei Ma and Ya-Qin Zhang and Lin Yan and Mu Qiao and Yong-Xu Wu and Mingxuan Wang},
  journal={ArXiv},
  year={2025},
  volume={abs/2503.14476},
  url={https://api.semanticscholar.org/CorpusID:277104124}
}

@article{Zhang2023AppAgentMA,
  title={AppAgent: Multimodal Agents as Smartphone Users},
  author={China. Xiaoyan Zhang and Zhao Yang and Jiaxuan Liu and Yucheng Han and Xin Chen and Zebiao Huang and Bin Fu and Gang Yu},
  journal={ArXiv},
  year={2023},
  volume={abs/2312.13771},
  url={https://api.semanticscholar.org/CorpusID:266435868}
}

@article{Furuta2023MultimodalWN,
  title={Multimodal Web Navigation with Instruction-Finetuned Foundation Models},
  author={Hiroki Furuta and Ofir Nachum and Kuang-Huei Lee and Yutaka Matsuo and Shixiang Shane Gu and Izzeddin Gur},
  journal={ArXiv},
  year={2023},
  volume={abs/2305.11854},
  url={https://api.semanticscholar.org/CorpusID:258823350}
}

@article{Wang2023VoyagerAO,
  title={Voyager: An Open-Ended Embodied Agent with Large Language Models},
  author={Guanzhi Wang and Yuqi Xie and Yunfan Jiang and Ajay Mandlekar and Chaowei Xiao and Yuke Zhu and Linxi (Jim) Fan and Anima Anandkumar},
  journal={Trans. Mach. Learn. Res.},
  year={2023},
  volume={2024},
  url={https://api.semanticscholar.org/CorpusID:258887849}
}

@article{Schick2023ToolformerLM,
  title={Toolformer: Language Models Can Teach Themselves to Use Tools},
  author={Timo Schick and Jane Dwivedi-Yu and Roberto Dess{\`i} and Roberta Raileanu and Maria Lomeli and Luke Zettlemoyer and Nicola Cancedda and Thomas Scialom},
  journal={ArXiv},
  year={2023},
  volume={abs/2302.04761},
  url={https://api.semanticscholar.org/CorpusID:256697342}
}

@article{Li2024EmbodiedAI,
  title={Embodied Agent Interface: Benchmarking LLMs for Embodied Decision Making},
  author={Manling Li and Shiyu Zhao and Qineng Wang and Kangrui Wang and Yu Zhou and Sanjana Srivastava and Cem Gokmen and Tony Lee and Li Erran Li and Ruohan Zhang and Weiyu Liu and Percy Liang and Fei-Fei Li and Jiayuan Mao and Jiajun Wu},
  journal={ArXiv},
  year={2024},
  volume={abs/2410.07166},
  url={https://api.semanticscholar.org/CorpusID:273234205}
}

@inproceedings{Narasimhan2015LanguageUF,
  title={Language Understanding for Text-based Games using Deep Reinforcement Learning},
  author={Karthik Narasimhan and Tejas D. Kulkarni and Regina Barzilay},
  booktitle={Conference on Empirical Methods in Natural Language Processing},
  year={2015},
  url={https://api.semanticscholar.org/CorpusID:8395799}
}

@article{Gur2023ARW,
  title={A Real-World WebAgent with Planning, Long Context Understanding, and Program Synthesis},
  author={Izzeddin Gur and Hiroki Furuta and Austin Huang and Mustafa Safdari and Yutaka Matsuo and Douglas Eck and Aleksandra Faust},
  journal={ArXiv},
  year={2023},
  volume={abs/2307.12856},
  url={https://api.semanticscholar.org/CorpusID:260126067}
}

@article{Yao2022ReActSR,
  title={ReAct: Synergizing Reasoning and Acting in Language Models},
  author={Shunyu Yao and Jeffrey Zhao and Dian Yu and Nan Du and Izhak Shafran and Karthik Narasimhan and Yuan Cao},
  journal={ArXiv},
  year={2022},
  volume={abs/2210.03629},
  url={https://api.semanticscholar.org/CorpusID:252762395}
}

@article{Hong2023CogAgentAV,
  title={CogAgent: A Visual Language Model for GUI Agents},
  author={Wenyi Hong and Weihan Wang and Qingsong Lv and Jiazheng Xu and Wenmeng Yu and Junhui Ji and Yan Wang and Zihan Wang and Yuxiao Dong and Ming Ding and Jie Tang},
  journal={2024 IEEE/CVF Conference on Computer Vision and Pattern Recognition (CVPR)},
  year={2023},
  pages={14281-14290},
  url={https://api.semanticscholar.org/CorpusID:266210390}
}

@article{Brohan2023RT2VM,
  title={RT-2: Vision-Language-Action Models Transfer Web Knowledge to Robotic Control},
  author={Anthony Brohan and Noah Brown and Justice Carbajal and Yevgen Chebotar and Krzysztof Choromanski and Tianli Ding and Danny Driess and Kumar Avinava Dubey and Chelsea Finn and Peter R. Florence and Chuyuan Fu and Montse Gonzalez Arenas and Keerthana Gopalakrishnan and Kehang Han and Karol Hausman and Alexander Herzog and Jasmine Hsu and Brian Ichter and Alex Irpan and Nikhil J. Joshi and Ryan C. Julian and Dmitry Kalashnikov and Yuheng Kuang and Isabel Leal and Sergey Levine and Henryk Michalewski and Igor Mordatch and Karl Pertsch and Kanishka Rao and Krista Reymann and Michael S. Ryoo and Grecia Salazar and Pannag R. Sanketi and Pierre Sermanet and Jaspiar Singh and Anikait Singh and Radu Soricut and Huong Tran and Vincent Vanhoucke and Quan Ho Vuong and Ayzaan Wahid and Stefan Welker and Paul Wohlhart and Ted Xiao and Tianhe Yu and Brianna Zitkovich},
  journal={ArXiv},
  year={2023},
  volume={abs/2307.15818},
  url={https://api.semanticscholar.org/CorpusID:260293142}
}

@inproceedings{Zhang2024CodeAgentEC,
  title={CodeAgent: Enhancing Code Generation with Tool-Integrated Agent Systems for Real-World Repo-level Coding Challenges},
  author={Kechi Zhang and Jia Li and Ge Li and Xianjie Shi and Zhi Jin},
  booktitle={Annual Meeting of the Association for Computational Linguistics},
  year={2024},
  url={https://api.semanticscholar.org/CorpusID:266999556}
}

@article{Rawles2024AndroidWorldAD,
  title={AndroidWorld: A Dynamic Benchmarking Environment for Autonomous Agents},
  author={Christopher Rawles and Sarah Clinckemaillie and Yifan Chang and Jonathan Waltz and Gabrielle Lau and Marybeth Fair and Alice Li and Will Bishop and Wei Li and Folawiyo Campbell-Ajala and Daniel Toyama and Robert Berry and Divya Tyamagundlu and Timothy P. Lillicrap and Oriana Riva},
  journal={ArXiv},
  year={2024},
  volume={abs/2405.14573},
  url={https://api.semanticscholar.org/CorpusID:269982433}
}

@article{Shridhar2020ALFWorldAT,
  title={ALFWorld: Aligning Text and Embodied Environments for Interactive Learning},
  author={Mohit Shridhar and Xingdi Yuan and Marc-Alexandre C{\^o}t{\'e} and Yonatan Bisk and Adam Trischler and Matthew J. Hausknecht},
  journal={ArXiv},
  year={2020},
  volume={abs/2010.03768},
  url={https://api.semanticscholar.org/CorpusID:222208810}
}

\newpage
\appendix

\section{Misaligned Learning Signals in OSWorld and AndroidWorld.}
\label{app:case}
\begin{figure*}[t]
    \centering
    \begin{subfigure}[b]{\linewidth}
\includegraphics[width=0.9\linewidth]{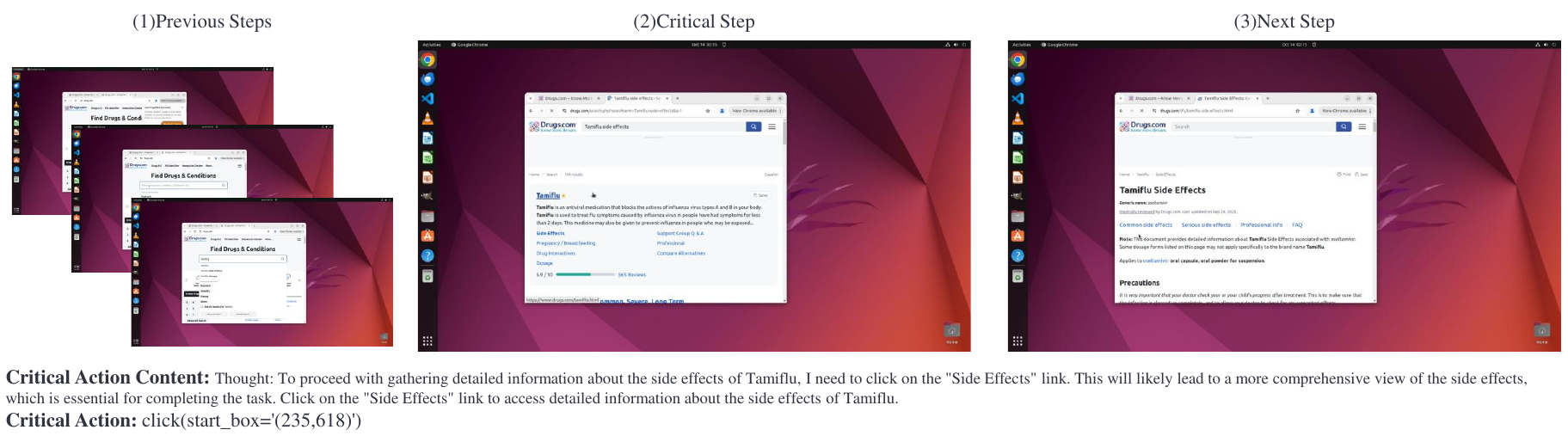}
    \caption{Successful Trajectory}
    \end{subfigure}
    \begin{subfigure}[b]{\linewidth}
    \includegraphics[width=0.9\linewidth]{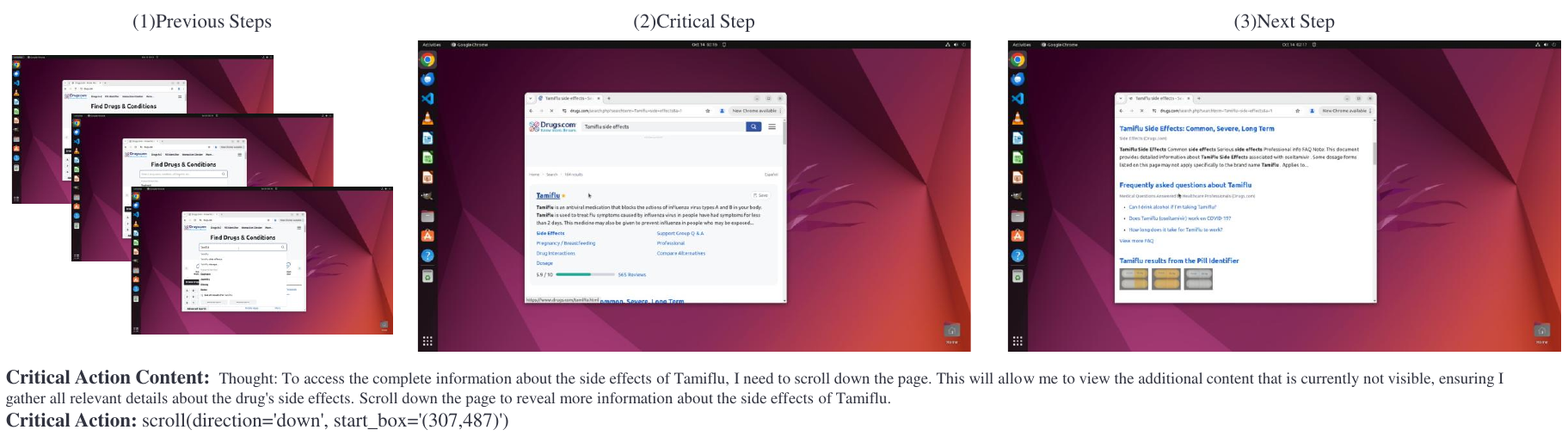}
    \caption{Failed Trajectory}
    \end{subfigure}
    \caption{
Comparison of success and failure trajectories in OSWorld for the task ``Show side effects of Tamiflu.''
}
    \label{fig:app-cases-os}
\end{figure*}

\begin{figure}[ht]
    \centering
    \begin{subfigure}[b]{0.9\linewidth}
\includegraphics[width=\linewidth]{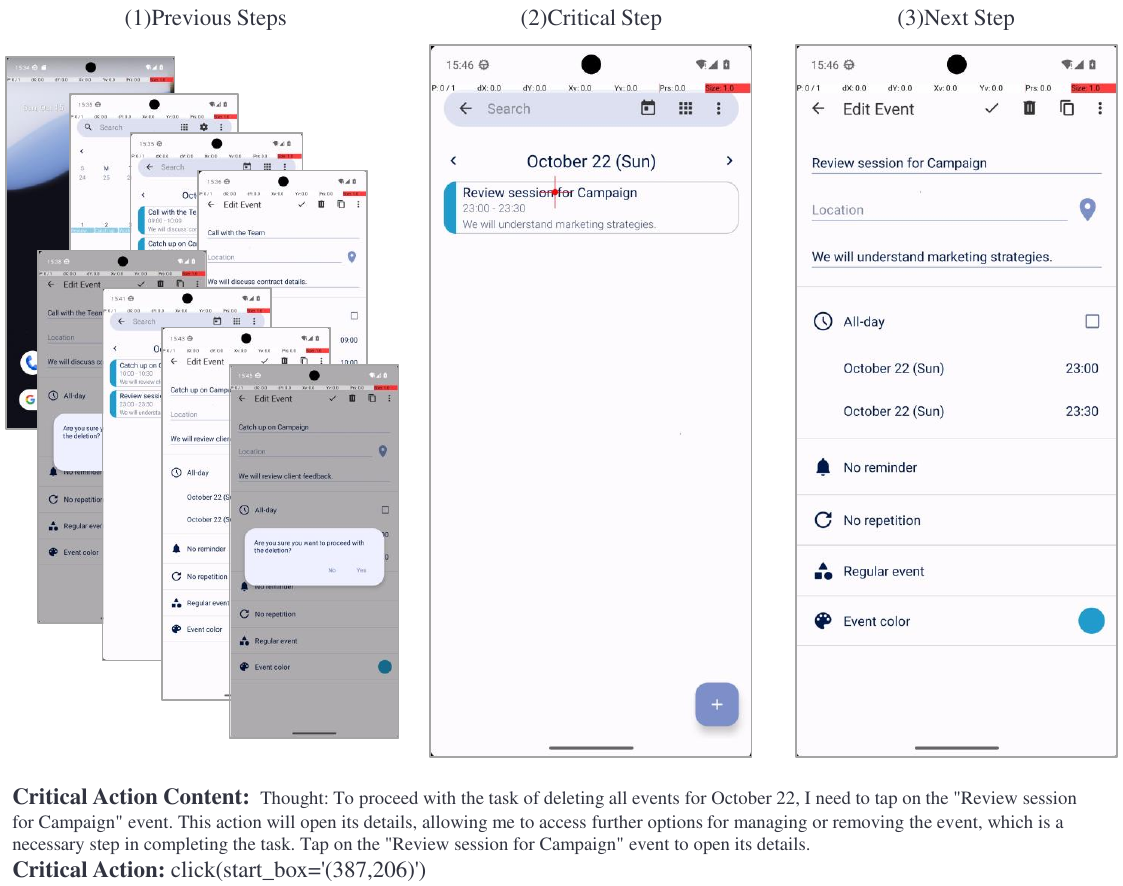}

    \caption{Successful Trajectory}
    \end{subfigure}
    \begin{subfigure}[b]{0.9\linewidth}
\includegraphics[width=\linewidth]{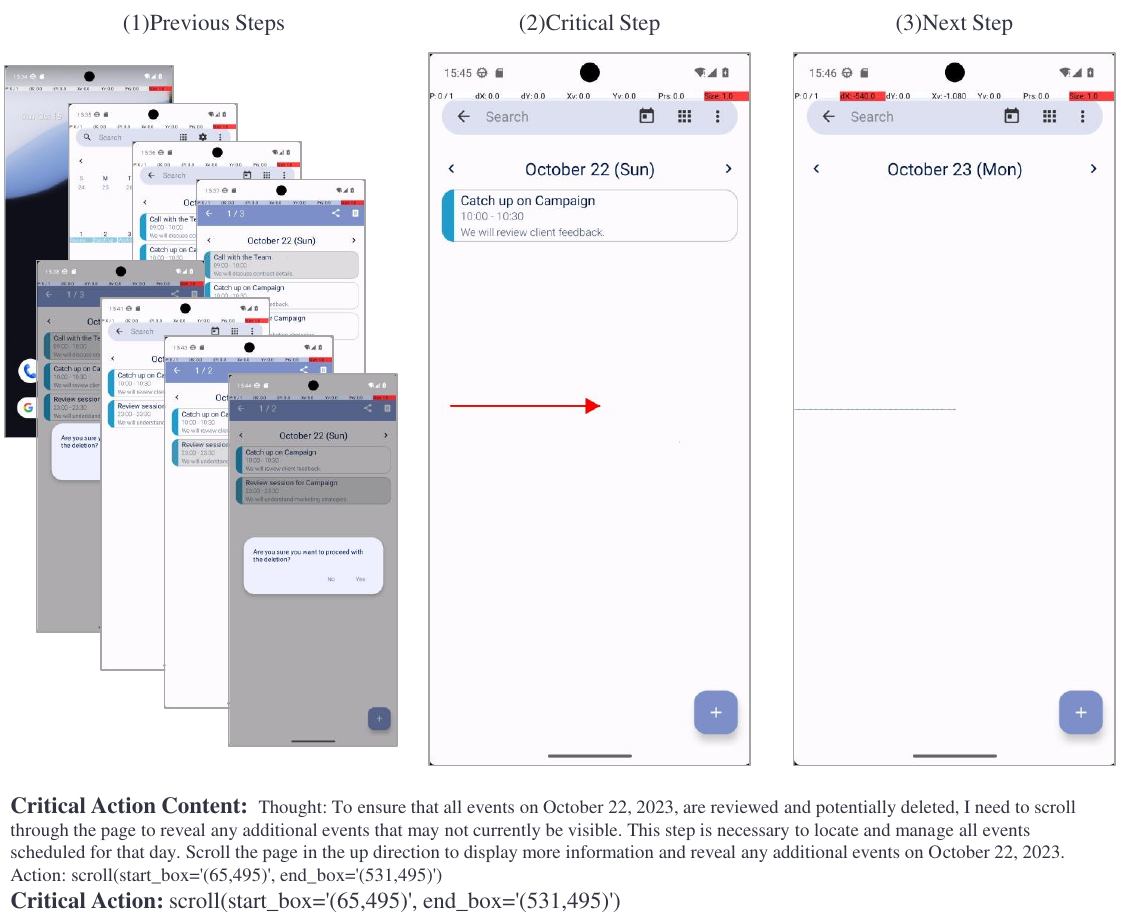}
    \caption{Failed Trajectory}
    \end{subfigure}
    \caption{
    Comparison of success and failure trajectories in AndoridWorld for the task ``In Simple Calendar Pro, delete all the calendar events on 2023-10-22.''}
    \label{fig:app-cases-aw}
\end{figure}

We present cases of misaligned learning signals in both OSWorld and AndroidWorld. As shown in Figure~\ref{fig:app-cases-os} and Figure~\ref{fig:app-cases-aw}, successful and failed trajectories for the same task often share similar intermediate reasoning steps, but diverge at a critical decision point where the failed trajectory takes an incorrect action. This example illustrates that failure trajectories, although their final outcome is negative, often contain many valid intermediate steps. Penalizing the entire trajectory can lead to incorrect learning for these correct steps.

\section{Setting}
\label{app:setting}
We provide our detail settings in Table~\ref{tab:all-setting} and Table~\ref{tab:setting}. 
\begin{table}[ht]
\centering
\resizebox{0.8\linewidth}{!}{
\begin{tabular}{lc}
\toprule
Hyperparameters & All methods\\
\midrule
Train batch size & 16 \\
PPO batch size & 256 \\
Training epoches & 8 \\
Rollout numbers  & 16   \\
Image tokens & 1350 \\
Temperature & 0.7 \\
Learning rate & 1e-6\\
Kl coefficient & 0.001 \\
GPU numbers & 16 \\
\bottomrule
\end{tabular}}
\caption{Model configurations for all methods.}
\label{tab:all-setting}
\end{table}

\begin{table}[ht]
\centering
\resizebox{\linewidth}{!}{
\begin{tabular}{lcc}
\toprule
Hyperparameters & T-GRPO & GiGRPO\&\name \\
\midrule
Max turns (OSWorld) & 20 & 30 \\
Max turns (AndroidWorld) & 20 & 25\\
Prompt length & 3076 & 4096   \\
Response length & 29692 (256 per turn) & 256 \\

\bottomrule
\end{tabular}}
\caption{Model configurations across methods.}
\label{tab:setting}
\end{table}

\end{document}